\documentclass[runningheads]{llncs}
\usepackage{graphicx}

\usepackage{tikz}
\usepackage{comment}
\usepackage{amsmath,amssymb} 
\usepackage{color}

\usepackage[accsupp]{axessibility}  

\usepackage{algorithm}
\usepackage[noend]{algpseudocode}
\usepackage{graphicx}
\usepackage{booktabs}
\usepackage{mathtools}
\usepackage{multirow}
\usepackage{arydshln}
\usepackage[normalem]{ulem}
\usepackage{comment}
\usepackage{subfig}

\renewcommand{\epsilon}{\varepsilon}
\renewcommand{\phi}{\varphi}

\DeclareMathOperator{\argmax}{\arg\,\max}

\newcommand\numberthis{\addtocounter{equation}{1}\tag{\theequation}}

\newcommand{\calQ}{\mathcal{Q}}

\newcommand{\calD}{\mathcal{D}}
\newcommand{\calL}{\mathcal{L}}

\newcommand{\calP}{\mathcal{P}}

\newcommand{\calS}{\mathcal{S}}

\newcommand{\bbR}{\mathbb{R}}
\newcommand{\bbI}{\mathbb{I}}

\newcommand{\bp}{\mathbf{p}}

\newcommand{\bz}{\mathbf{z}}

\newcommand{\feat}{\mathbf{f}}

\newcommand{\vid}{\ensuremath{\text{V}}}

\newcommand{\xobs}{\ensuremath{\mathbf{\text{X}}^{ob}}}
\newcommand{\zmiss}{\ensuremath{\mathbf{\text{X}}^{ms}}}
\newcommand{\param}{\ensuremath{\mathbf{\Theta}}}

\newcommand{\PP}{\ensuremath{\mathbb{P}}}

\newcommand{\EE}{\ensuremath{\mathbb{E}}}

\newcommand{\BK}[1]{ {\left( #1 \right)} }

\newcommand{\singleframe}{Timestamp Supervision}
\newcommand{\randomframe}{SkipTag Supervision}

\makeatletter
\DeclareRobustCommand\onedot{\futurelet\@let@token\@onedot}
\def\@onedot{\ifx\@let@token.\else.\null\fi\xspace}

\makeatother

\begin{document}
\pagestyle{headings}
\mainmatter
\def\ECCVSubNumber{4788}  

\title{A Generalized \& Robust Framework For Timestamp Supervision in\\
Temporal Action Segmentation} 


\titlerunning{Robust Timestamp Supervision}
%
\author{Rahul Rahaman\inst{1} 
Dipika Singhania\inst{2} 
Alexandre Thiery\inst{3} 
Angela Yao\inst{4} 
}
\authorrunning{R. Rahaman et al.}
\institute{
\email{rahul.rahaman@u.nus.edu, dipika16@comp.nus.edu.sg, a.h.thiery@nus.edu.sg, ayao@comp.nus.edu.sg}\\
National University of Singapore
}
\maketitle

\begin{abstract}
    In temporal action segmentation, \singleframe{} requires only a handful of labelled frames per video sequence. For unlabelled frames, previous works rely on assigning hard labels, and performance rapidly collapses under subtle violations of the annotation assumptions. We propose a novel Expectation-Maximization (EM) based approach that leverages the label uncertainty of unlabelled frames and is robust enough to accommodate possible annotation errors. 
    With accurate timestamp annotations, our proposed method produces SOTA results and even exceeds the fully-supervised setup in several metrics and datasets. When applied to timestamp annotations with missing action segments, our method presents stable performance. To further test our formulation's robustness, we introduce the new challenging annotation setup of \randomframe{}. This setup relaxes constraints and requires annotations of any fixed number of random frames in a video, making it more flexible than \singleframe{} while remaining competitive. 
\end{abstract}

\section{Introduction}
\label{sec:intro}

\begin{figure}[t!]
\begin{center}
\includegraphics[width=1.0\linewidth]{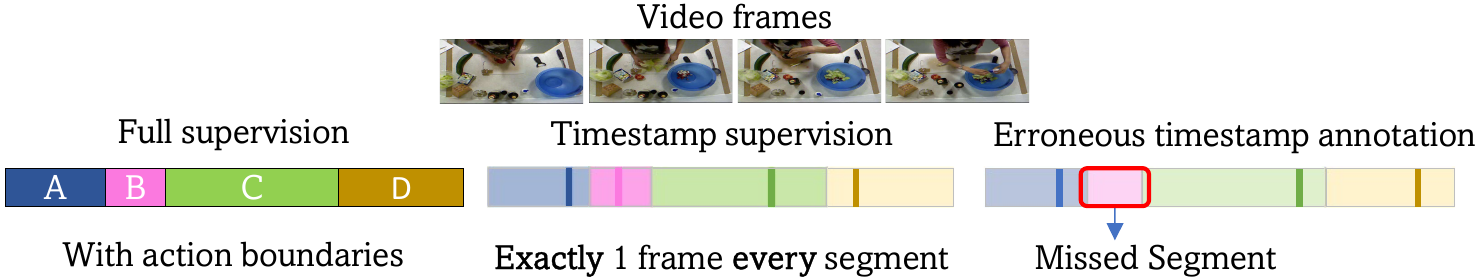}
\end{center}
\vspace{-4mm}
\caption{\textbf{Annotations levels.} Full-supervision requires labels for every frame in the video. \singleframe{} requires labels for a \textit{single} frame from \textit{each} action segment, although annotators may miss some action segments.}
\label{fig:levels-of-supervision}
\vspace{-2mm}
\end{figure}

Temporal action segmentation partitions and classifies a sequence of actions in long untrimmed video.
In a fully-supervised setting, the exact action boundaries and actions of every frame are labelled for all training videos. \textbf{\singleframe{} (TSS)}~\cite{SingleFrameDimaDamen,SingleFrameJGall} is a lightweight alternative in which the annotator labels one frame from each action in the video (Fig \ref{fig:levels-of-supervision}). 
TSS requires magnitudes fewer labels than the fully-supervised setting\footnote{On Breakfast Actions~\cite{kuehne2014language}, only 0.032\% of the frames need labels!}, but it has a strict constraint -- there must be a timestamp for \textbf{every action} in the video sequence. To fulfill this constraint, annotators must watch the entire video carefully, so as not to miss any action segments. This affects the annotation efficiency 
and leads to only a 35\% reduction in annotation time compared to full supervision. Furthermore, annotators may \textit{mistakenly} skip actions, as observed in our user study on timestamp labelling.  The annotators' slow performance and mistakes highlight the impracticality of the constraints of \singleframe{}.


 Existing works on TSS~\cite{SingleFrameDimaDamen,SingleFrameJGall} are based on strong assumptions.  For example, \cite{SingleFrameDimaDamen} directly parameterizes the position and width of action segments around the annotated timestamps, while \cite{SingleFrameJGall} sets one action boundary with a handcrafted energy function between two annotated frames and assigns hard labels to in-between unlabelled frames. The performance of these methods collapses when annotation errors are introduced (see Fig.~\ref{fig:perf-vs-error}) as the errors violate the condition ``\textit{one} timestamp for \textit{every} action segment''. Note that this sort of annotation error differs from the typical annotation error of mislabelling as it emerges from the difficulty of adhering to annotation constraints.

In this work, we first propose a novel, general model formulation for \singleframe{}, called \textbf{EM-TSS}. EM-TSS uses the first principles of an Expectation-Maximization (E-M) formulation to model 
frame-wise labels conditioned on action boundary locations. We restructure the EM formulation for the temporal segmentation task so that the maximization step reduces to a frame-wise cross-entropy minimization. Unlike the heuristics of~\cite{SingleFrameDimaDamen,SingleFrameJGall}, EM-TSS has theoretical groundings and assumes only an extremely weak prior distribution on segment length. EM-TSS also makes no additional assumptions per action segment and can leverage the label uncertainty provided by the E-M formulation to avoid assigning hard labels. EM-TSS surpasses previous works~\cite{SingleFrameDimaDamen,SingleFrameJGall} for the majority of metrics by a significant margin. For several datasets and metrics, 
it even outperforms the fully-supervised setting with only a handful of labels. We attribute the success to the use of more accurate (albeit fewer) labels and verify with a user study and ablations on the position of the timestamps.

Next, we generalize our EM framework into \textbf{EM-Gen} to handle 
errors in \singleframe{} due to annotators missing action segments between timestamps. EM-Gen's performance remains stable under such annotation error and can tolerate up to 20\% missing segments, with a marginal drop in performance compared to the TSS methods. To further challenge the performance limits of EM-Gen, we introduce a new \textit{SkipTag supervision} to allow for random frame annotations anywhere in the video. SkipTag supervision is far more flexible than timestamp supervision and is much less restrictive. Applying EM-Gen to SkipTag produces good results despite this weaker form of annotation.
 
Summarizing the \textbf{contributions} in this work, 
\begin{enumerate}
    \item We propose a novel E-M based method, EM-TSS, for action segmentation under \singleframe{}. EM-TSS surpasses SOTA timestamp methods by a large margin and is competitive with fully-supervised methods. 
    \item We generalize EM-TSS into EM-Gen, which, unlike competing works, remains robust under subtle timestamp annotation errors.
    \item We push the limits of EM-Gen and apply it to the weaker SkipTag supervision, which enables the free annotation of a random selection of frames.
    \item 
    Through a user-study, we compare the annotation efforts for Timestamp and SkipTag supervision and investigate the ambiguity of boundary annotations. 
\end{enumerate}

\section{Related Works}
\label{sec:related-works}

\noindent \textbf{Temporal Action Segmentation} assigns frame-wise action labels to long, untrimmed video; it is a 1-D temporal analogue of semantic segmentation. Recent approaches~\cite{li2020ms,sener2020temporal,wang2020boundary,ishikawa2021alleviating,chen2020action,wang2020gated,singhania2021coarse,gao2021global2local} have used pre-extracted I3D~\cite{trimmed2d-simonyan2014two} snippet-level features as input and modelled the temporal structure using various models, such as HMMs~\cite{kuehne2014language},  RNNs~\cite{singh2016multi} or TCNs~\cite{TED-lea2017temporal,wang2020gated,li2020ms,chen2020action}. Temporal Convolution Networks (TCNs)~\cite{TED-lea2017temporal,li2020ms,farha2019ms} are shown to be the most effective, with the popular SOTA being MSTCNs~\cite{li2020ms}. We use I3D features as input and the MSTCN architecture as a backbone, similar to previous TSS work~\cite{SingleFrameJGall}.

\textbf{Fully-Supervised Methods}~\cite{TED-lea2017temporal,farha2019ms,kuehne2014language} 
require action labels to be provided for all frames. To reduce the annotation costs, \textbf{weakly supervised} methods only require an ordered~\cite{TED-ding2018weakly,weakly-richard2018neuralnetwork,weakly-chang2019d3tw,weakly-li2019weakly,weakly-souri2019fast,Rashid_2020_WACV} or unordered list of actions~\cite{richard2018action,Li_2020_CVPR,Fayyaz_2020_CVPR,ding2021temporal} as supervision. \textbf{Unsupervised} approaches~\cite{unsupervised-sener2018unsupervised,unsupervised-kukleva2019unsupervised,unsupervised-vidalmata2021joint} do not require any annotation and are primarily focused on clustering segments. 

\textbf{Timestamp Supervision (TSS)} has recently emerged as a new form of weak supervision that requires a single frame annotation for each action~\cite{ju2021divide,prevsegmentrnn-kuehne2018hybrid,SingleFrameDimaDamen,SingleFrameJGall}. 
TSS thus requires segmentation methods to utilize and estimate a huge amount of unlabelled training frames. TSS is the preferred choice over other weak supervision methods due to its higher accuracy yet lower annotation cost~\cite{SingleFrameDimaDamen}.

Existing works ~\cite{SingleFrameDimaDamen,SFNetMaEtal} expand action segments' length around the annotated frames based on discriminative probability outputs.
Li et al.~\cite{SingleFrameJGall} fix a boundary between two consecutive annotated frames based on similarity measures and assign the frames in between to either of the two actions. 
Our work emphasizes boundary localization. However, unlike solutions using hard boundaries~\cite{SingleFrameJGall} or distributional assumption \cite{SingleFrameDimaDamen,SingleFrameJGall}, 
we formulate boundary estimation and annotate unlabelled frames in the training videos with the principles of the E-M algorithm. Our problem formulation and derivation of the steps for the E-M algorithm stem naturally from the model assumptions of the supervised setup.

\section{Preliminaries}
\label{sec:background}

\subsection{Temporal Action Segmentation Task}\label{subsec:bgactionseg}
\noindent Consider a video $\vid$ with duration $T$. For fully-supervised action segmentation, each frame $\vid[t]: t \le T$ is labelled with the ground truth action label $y[t] \in \{1, \dots, C \}$ of $C$ possible action classes. The standard practice in action segmentation~\cite{li2020ms,wang2020boundary,singhania2021coarse} is to use pre-trained frame-wise features,~e.g.,~I3D \cite{carreira2017I3D} features. 
We denote these features as $\feat[t] \in \bbR^{d}$. The TCN takes $\feat$ as input and generates frame-wise action class probabilities $\bp$. For frame $t$, the vector $\bp[t]$ is a $C$-dimensional vector and $p_{t,c} := \bp[t,c]$ is the probability value assigned to action class $c$. A cross-entropy loss can be used to train the model.
\begin{equation}\label{eq:fully-sup-CE-loss}
    \calL_{\text{CE}} := - \sum_{t=1}^T \sum_{c=1}^C \omega_{t,c} \cdot \log p_{t,c}
\end{equation}
Under full supervision, $\omega$ can be set as the ground truth labels,~i.e.,~$\omega_{t,c} = \bbI\big[y[t] = c\big]$. To use the same loss under \singleframe{}, however, the weights $\omega_{t,c}$ must be estimated for the unlabelled frames where $y[t]$ is unknown. Previous works~\cite{SingleFrameDimaDamen,SingleFrameJGall} do so by setting hard weights,~i.e.,~$\omega_{t,c} \in \{0,1\}$, through some rules or a ranking scheme.

\subsection{E-M Algorithm} 
\label{subsec:em-algo}
\noindent Expectation-Maximization (E-M)~\cite{EMAlgo} is a classic iterative method for maximum likelihood estimation (MLE) with hidden variables and/or missing data. We consider the classical form with missing data.
Formally, let $\xobs, \zmiss$ be the observed and missing parts of the data, respectively, and $\calP(\xobs, \zmiss| \param)$ be the data likelihood conditioned on some unknown parameter \param{},~e.g.,~ the weights of a neural network model. 
As \xobs{} is observed and \zmiss{} is missing, directly maximizing $\calP$ as a function of \param{}, and hence obtaining the MLE of $\param^*$, is not possible. The E-M algorithm provides an iterative approach to obtain the MLE of \param{} under this setting through the use of a $\calQ$-function.
In simple terms, the $\calQ$-function serves as an expected version of the log-likelihood $\calP$ by replacing terms involving unobserved \zmiss{} with the expected value based on our current estimate of the parameter. Formally, let $\param^{(m)}$ be the current estimate of \param{} after the $m^{th}$ iteration. At iteration $(m+1)$, the \textit{Expectation step} calculates the $\calQ$-function:
\begin{gather}\label{eq:EMexpect}
    \calQ(\param{}, \param^{(m)}) = \EE_{\bz} \big[\log \calP(\xobs, \bz\,|\, \param)\big],
\end{gather}
where $\bz \sim f(\zmiss \,|\, \xobs, \param^{(m)})$,~i.e.,~the posterior distribution of the unobserved data $\zmiss$ given the observed data $\xobs$ and the current estimate of the parameters $\param^{(m)}$. 
Note that the term in Eq. \ref{eq:EMexpect} is a function of the parameters \param{}. In the \textit{Maximization step}, the $\calQ$ function is then maximized with respect to the parameters \param{} to obtain the new estimate 
\begin{gather}\label{eq:EMmaxi}
    \param^{(m+1)} = \underset{\param{}}{\argmax} \; \calQ(\param{}, \param^{(m)}).
\end{gather}
This process is repeated until convergence,~i.e.,~$\param^{(m+1)}$ is desirably close to $\param^{(m)}$, or for a fixed number of iterations. The model (TCN) parameter learning happens during this M-step (Maximization), whereas the E-step (Expectation) iteratively updates the criteria to be optimized in the M-step.

\section{Method}
\label{sec:method}

\noindent 
The assignment of hard labels to each frame may lead to some ambiguities at the action boundaries and deteriorate performance (see Sec.~\ref{subsec:analysis-annotation}). To avoid the pitfalls of hard labels, we allow soft weights for $\omega$ in Eq.~\ref{eq:fully-sup-CE-loss}. We arrive at these soft weights using the E-M framework, as we show for \singleframe{} (Sec.~\ref{subsec:singleframe}), and extend it to the case of missing timestamps (Sec.~\ref{subsec:random-frame}).

\begin{figure}[t!]
\begin{center}
\includegraphics[width=0.6\linewidth]{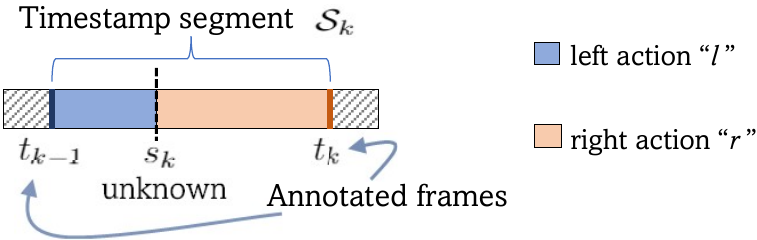}
\end{center}
\vspace{-4.0mm}
\caption{Depiction of \textbf{timestamp segment} $\calS_k$.
}
\label{fig:single-frame-segment}
\vspace{-4.0mm}
\end{figure}

\subsection{Segment-Based Notation}
\label{subsec:segnotation}
\noindent While a video is typically treated as a sequence of frames (Sec.~\ref{subsec:bgactionseg}), it can also be regarded as a sequence of action segments. Each segment is a set of contiguous frames belonging to one action, 
represented by a single color in Fig.~\ref{fig:levels-of-supervision}. 
A video with $K$ action segments will have action classes $c_1,\ldots,c_K$ out of $C$ action classes. $K$ may vary for different videos; the same action may also occur multiple times in a video. As the adjacent action segments share boundaries, $c_k \neq c_{k+1}$ for all $k < K$. We denote $s_k$ to be the starting frame for action segment $k$, with $s_1 \equiv 1$ and $s_k < s_{k+1}\le T$. Frame $s_k$ is also the \textit{boundary} between action segments $k\!-\!1$ and $k$. In the fully-supervised setup, every frame is labelled; this is equivalent to labelling start frame $s_k$ and action class $c_k$ for each segment $k$. 

\subsection{Timestamp Supervision}
\label{subsec:singleframe}

\noindent \singleframe{} provides one labelled frame or \textit{`timestamp'} per action segment, resulting in $K$ labelled frames $t_1,\ldots,t_K$ and action labels $c_1,\ldots,c_K$. There is exactly one timestamp, randomly positioned within the segment, for \emph{every} action segment,~i.e.,~$s_{k}\!\le\!t_{k}\!<\!s_{k+1}$, or equivalently $t_{k-1}\!<\!s_{k}\!\le\!t_{k}, \forall k$. As the action boundaries are unknown, 
we additionally define a \emph{timestamp segment}; the $k^{th}$ timestamp segment is denoted by $\calS_k$ and is bounded by frames $[t_{k-1}, t_{k}-1]$ (see Fig.\ref{fig:single-frame-segment}). Within $\calS_k$, the label $l := y[t_{k-1}] = c_{k-1}$ denotes the \textit{`left action'} class and $r := y[t_{k}] = c_k$ denotes the \textit{`right action'} class\footnote{We do not discuss the first and last timestamp segments $\calS_1:=[1,t_1)$ and $\calS_{K+1}:=[t_K, T]$ as their action labels $c_1, c_K$ are known.}. \\

\noindent \textbf{Timestamp Segment Likelihood:}
If $s_{k}$ is known, all the frames $[t_{k-1},s_k)$ can be labelled with action $l$ and frames $[s_{k},t_{k})$ can be labelled with action $r$. Hence, the likelihood of segment $\calS_k$ given $s_{k}\!=\!j$, denoted as $\calP_j(\calS_k \,|\, \param)$, becomes
\begin{gather} \label{eq:segmentGivenBoundary}
    \calP_j(\calS_k \,|\, \param) := \calP(\calS_k \;|\; s_{k}\!=\!j, \param) = \prod^{j-1}_{i=t_{k-1}} p_{i,l}^{\param} \prod^{t_{k}-1}_{i=j} p_{i,r}^{\param}.
\end{gather}
In Eq.~\ref{eq:segmentGivenBoundary}, we add a super-script $\param$ to $p_{t,c}^{\param}$ to emphasize the dependence of the probability $p_{t,c}$ on the model parameters $\param$ of the temporal convolutional network.  
If $s_{k}$ is known for each $k$, then maximizing the likelihood of Eq.~\ref{eq:segmentGivenBoundary} after taking the negative $\log$ and summing over all $K$ segments simplifies to minimizing the fully-supervised frame-wise cross-entropy (CE) loss of Eq.~\ref{eq:fully-sup-CE-loss}. Note that Eq.~\ref{eq:segmentGivenBoundary} and fully-supervised Eq.~\ref{eq:fully-sup-CE-loss} both assume independence of the frame-wise probability when conditioned on the network parameters and architecture.

In~\singleframe{}, the boundary $s_k$ is unknown, so we treat it as a random variable, with some prior probability of it being located at frame $j$,~i.e.,~$\pi_{k}(j) := \pi(s_{k}=j)$. The joint likelihood of segment $\calS_k$ and boundary $s_{k}$ then becomes $\calP(\calS_k, s_{k}=j\,|\, \param) = \calP_j(\calS_k\,|\,\param) \cdot \pi_k(j)$. We further elaborate on the prior distribution in subsection~\ref{subsec:prior}. 
As boundary $s_k$ can be located at any of the frames $[t_{k-1}, t_{k})$, the summarized joint likelihood can be written as
\begin{align} \label{eq:singleFrameFinalLikeli}
    \calP(\calS_k, s_{k}\,|\, \param) &= \prod_{j=t_{k-1}}^{t_{k}-1} \big[ \calP_j(\calS_k\,|\, \param) \cdot \pi_{k}(j) \big]^{\bbI[s_{k}=j]}.
\end{align}
\noindent \textbf{$\calQ$-Function \& Posterior.} 
Following Eq.~\ref{eq:EMexpect}, the $\calQ$-function for timestamp segment $\calS_k$ is obtained by taking the expectation of the log of the likelihood in Eq.~\ref{eq:singleFrameFinalLikeli} with respect to $f(s_{k} \,| \calD; \param^{(m)})$.  This represents the posterior distribution of boundary $s_{k}$ given observed data $\calD$ under the previous estimate of parameter $\param^{(m)}$. Here, observed data $\calD$ contains all the training information, such as input video features $\feat$ and the annotations $y[t_{k-1}], y[t_{k}]$. The $\calQ$-function for $\calS_k$, denoted by $\calQ_k(\param{}, \param^{(m)})$, thus becomes
\begin{align*}
    \calQ_k(\param{}, \param^{(m)})
    &=  \sum_{j=t_{k-1}}^{t_{k}-1}\!\! \EE_{s_{k}} \Big[\bbI[s_{k}\!=\!j] \cdot \log \Big( \calP_j(\calS_k\,|\, \param) \cdot \pi_{k}(j) \Big) \Big] \\  
    &= \Lambda + \sum_{j=t_{k-1}}^{t_{k}-1}\!\! \PP^{(m)}[s_{k}\!=\!j | \calD] \cdot \log \calP_j(\calS_k\,|\, \param), \numberthis{}\label{eq:qfunc-single-frame}
\end{align*}
where $\PP^{(m)}[s_{k}=j | \calD]$ is the posterior probability that $s_{k}$ is located at frame $j$ given the data and under the model parameter $\param^{(m)}$.
$\Lambda$ consists of the terms (prior $\pi_k(j)$) that are constant with respect to the parameter \param{} and hence can be ignored during the maximization of $\calQ$ with respect to \param{}.
The posterior probabilities can be obtained by
\begin{align}\label{eq:posteriorBoundarySF}
    \PP^{(m)}[s_{k}=j | \calD] = \frac{\calP_j(\calS_k \,|\, \param^{(m)}) \cdot \pi_{k}(j)}{ \sum_{i=t_{k-1}}^{t_{k}-1} \calP_i(\calS_k\,|\, \param^{(m)}) \cdot \pi_{k}(i)},
\end{align}
where $\calP_j(\calS_k\,|\, \param^{(m)})$ is defined in Eq.~\ref{eq:segmentGivenBoundary}. The final $\calQ$-function for the whole video is calculated by $\calQ(\param{}, \param^{(m)}) = \sum_k \calQ_k(\param{}, \param^{(m)})$. \\

\noindent\textbf{$\calQ$-function to Frame-Wise CE Loss.} 
The formulation of the $\calQ$-function in Eq.~\ref{eq:qfunc-single-frame} requires estimating two terms for each frame $j$. The first is $\PP^{(m)}[s_{k}=j | \calD]$, which can be calculated only once per E-step as it does not depend on $\param$. 
The second term $\calP_j(\calS_k\,|\, \param)$, however, must be calculated for every iteration of the M-step. 
Even after obtaining frame-wise probabilities $p_{j,c}^\param$ from the model, every calculation of $\log \calP_j(\calS_k | \param)$ for each frame $j$ (as given in Eq.~\ref{eq:segmentGivenBoundary}) takes linear time with respect to video length,~i.e.,~of complexity $\mathcal{O}(T)$ per frame $j$. This may make each epoch of the parameter learning in the M-step computationally expensive. To reduce the computational complexity, we simplify Eq.~\ref{eq:qfunc-single-frame} by substituting $\log \calP_j(\calS_k; \param)$ from Eq.~\ref{eq:segmentGivenBoundary} and rearranging the sums to get 
\begin{gather*}\label{eq:Qfunc-as-CE}
  \calQ_k(\param{}, \param^{(m)}) = \sum_{j=t_{k-1}}^{t_{k}-1} \,\sum_{c=1}^C \,\omega^{(m)}_{j,c} \cdot \log p_{j,c}^{\param}\,\,,\numberthis{}\\
  \text{where} \qquad  \omega^{(m)}_{j,c} = \left\{\begin{array}{ll}
                         \sum_{i=j+1}^{t_{k}-1} \PP^{(m)}[s_{k}=i | \calD] & : \,c=l \\
                         1 - \omega^{(m)}_{j,l} & : \,c=r \\
                         0 & : \, \text{otherwise.} \\ %
    \end{array}\right.
\end{gather*}

This new form of $\calQ_k(\param{}, \param^{(m)})$ is more intuitive than Eq.~\ref{eq:qfunc-single-frame} and, as desired at the start of this section, reduces to a negative frame-wise CE loss (Eq.~\ref{eq:fully-sup-CE-loss}). Furthermore, our posterior weights $\omega^{(m)}$ are not hard labels, and the weights capture the label uncertainty of the unlabelled frames. This simplified formulation converges in very few iterations and adds negligible differences to the training time compared to a fully-supervised approach.
The weights $(\omega^{(m)}_{j,l}, \omega^{(m)}_{j,r})$ are actually the posterior probabilities (given data and parameters) that frame $j$ belongs to the left and right actions $l$ and $r$, respectively. The weight $\omega^{(m)}_{j,l}$ acts as the current estimate of $\bbI\big[y[j] = l\big]$ (see Eq.~\ref{eq:fully-sup-CE-loss}) when $y[j]$ is unknown.
Furthermore, regardless of the current parameter $\param^{(m)}$, the weights have the property that $\omega^{(m)}_{t_{k-1},l}\!=\!1$ and $\omega^{(m)}_{j,l}\!\ge\!\omega^{(m)}_{j+1,l},\, \omega^{(m)}_{j,r}\!\le\!\omega^{(m)}_{j+1,r}$ for any $t_{k-1}\!\le\!j\!<\!t_{k}$. This means the weight assigned to left action class $l$ is $1$ for frame $t_{k-1}$ (as it should be), and it gradually decreases as $j$ increases, with the converse being true for the action class $r$. These weights can be calculated during the E-step and be used in a CE loss without additional computation during the M-step. \\

\noindent \textbf{Initialization and Overall Procedure.}
The E-M algorithm requires good initialization. Following~\cite{SingleFrameJGall}, we initialize the TCN parameters with $\param^{(0)}$ derived from a `\textit{Naive}' baseline learned with a frame-wise cross-entropy loss applied \textit{only} to the annotated frames $t_k$. From this initialization, we perform the first E-step to obtain the frame-wise weights $\omega^{(0)}_{j,c}$. Thereafter, we iterate between Maximization (maximize $\calQ$) and Expectation (update $\omega^{(m)}$) steps, which we call \textbf{EM-TSS}, to find the best estimate $\param^*$ (see the pseudo-code in algorithm~\ref{algo:em}).

\begin{algorithm}[h]
\caption{Our iterative procedure for EM-TSS}\label{algo:em}
\begin{algorithmic}[1]
\For {\emph{epoch} $\le N^{init}$} \Comment{\textbf{\emph{Initialization}}:}
    \State calculate $\calL := - \sum_{k=1}^K \log p^{\param}_{t_k,c_k}$
    \State minimize $\calL$, update $\param$
\EndFor
\State set $\param^{(0)} = \param$, set $m=0$
\For {$m < M$}
    \State calculate $\omega^{(m)}$ using $\param^{(m)}$ \Comment{\textbf{\emph{Expectation}}:}
    \For {\emph{epoch} $\le N^{max}$} \Comment{\textbf{\emph{Maximization}}:}
        \State maximize $\calQ(\param{}, \param^{(m)})$, update \param{}
    \EndFor
    \State set $\param^{(m+1)} = \param$
\EndFor
\State set final $\param^* = \param^{(M)}$
\end{algorithmic}
\end{algorithm}
\vspace{-4.0mm}

\subsection{Timestamp Supervision Under Missed Timestamps}
\label{subsec:random-frame}
\singleframe{} places a strong constraint that there is one timestamp for \textit{every} action segment present in the video. In practice, as revealed by our user study (see \textit{Supplementary}), annotators may miss action segments. These missed segments, as we show later in Sec~\ref{sec:experiments}, cause TSS methods to collapse rapidly.

\begin{figure}[b!]
\begin{center}
\includegraphics[width=0.8\linewidth]{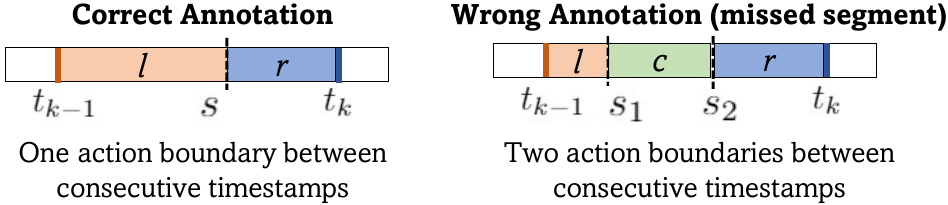}
\end{center}
\vspace{-2.0mm}
\caption{\textbf{Correct vs. Missed Segments.} Timestamp segment $\calS$ between consecutive timestamps $t_{k-1}, t_k$ with action labels $l, r$ respectively.  When correct, there is one unknown action boundary $s$; when a segment is missed, there are actually two unknown boundaries, $s_1$ and $s_2$, and an unknown action $c$ in between.}
\label{fig:random-frame-segment}
\end{figure}

As such, we target the impracticality of having perfectly accurate timestamps and develop a method that is more robust to such annotation errors. Specifically, we generalize our E-M framework \textbf{EM-TSS} into \textbf{EM-Gen} to accommodate possible missing segments between two consecutive annotated timestamps, as depicted in Fig.~\ref{fig:random-frame-segment}. The overall E-M framework remains the same as Timestamp supervision, but differs in the derivation of the associated $\calQ$-function and weights $\omega$. We consider the timestamp segment $[t_{k-1}, t_k]$ with the following cases .


\textbf{Case 1 ($C_1$): Correct timestamp annotation.} In this case, there will be exactly \textit{one action boundary} $s$, where $t_{k-1}\!<\!s\!\le\!t_{k}$ and $r\neq l$.  
This case is the same as a typical timestamp segment of TSS; given boundary location $s$, the segment likelihood is the same as Eq. \ref{eq:segmentGivenBoundary} earlier,
\begin{equation*}
    \calP(\calS | C_1,s, \param) = \prod^{s-1}_{i=t_{k-1}} p_{i,l}^{\param} \prod^{t_{k}-1}_{i=s} p_{i,r}^{\param}.
\end{equation*}
\textbf{Case 2 ($C_2$): Missed timestamp annotation.} In this case, there will be \textit{two action boundaries} $s_1, s_2$, which  define a middle action segment such that $t_{k-1}\!<\!s_1\!<\!s_2\!\le\!t_{k}$. 
If the middle action label is $c$, where $c \neq l$ and $c \neq r$  (this allows for both $r\!\neq\!l$ and $r\!=\!l$), the segment likelihood is 
\begin{equation*}
    \calP(\calS | C_2,s_1,s_2,c, \param) = \prod^{s_1-1}_{i=t_{k-1}} p_{i,l}^{\param} \,\prod^{s_2-1}_{i=s_1} p_{i,c}^{\param} \prod^{t_{k}-1}_{i=s_2} p_{i,r}^{\param}.
\end{equation*}

Similar to EM-TSS, to calculate the $\calQ$-function, we now need to combine the log-likelihoods with the corresponding weights as the posterior probabilities. The $\calQ$-function for the segment $\calS$ is calculated as
\begin{align*}
    \calQ(\param, \param^{(m)}) = \sum\nolimits_s &\PP^{(m)}[C_1, s | \calD] \cdot \log \calP(\calS | C_1,s, \param) \,\, + \\
    \sum\nolimits_{\tau} &\PP^{(m)}[C_2, \mathbf{\tau} | \calD] \cdot \log \calP(\calS | C_2,\tau, \param), \numberthis{}\label{eq:Qfunc-randomframe}
\end{align*}
where $\tau$ indexes all valid triplets $(s_1, s_2, c)$ of Case 2.
In the first term, $\PP^{(m)}[C_2, s | \calD]$ denotes the posterior probability that the segment belongs to Case 1 \textit{and} the action boundary is located at $s$, given the data and under parameter $\param^{(m)}$. The second term contains an analogous posterior for Case 2 and some triplet $\tau$. 
The probabilities of Eq.~\ref{eq:Qfunc-randomframe} can be estimated in the same way as Eq.~\ref{eq:posteriorBoundarySF} of EM-TSS. A few components in the $\calQ$-function have high computational complexity, e.g., the second sum in Eq.~\ref{eq:Qfunc-randomframe} is of $\mathcal{O}(T^2)$ where $T$ is the video duration. However, a few rearrangements similar to the EM-TSS change the $\calQ$-function into the same form as in Eq.~\ref{eq:Qfunc-as-CE}. We defer the exact forms to the \textit{Supplementary}.
The final posterior weights become 
\begin{align*}
    \omega^{(m)}_{j,l} &= \sum_{s>j} \PP^{(m)}[C_1, s | \calD] + \sum_{\tau:s_1 > j}\PP^{(m)}[C_2, \tau | \calD]\\
    \omega^{(m)}_{j,c} &= \sum_{s_1\le j} \sum_{s_2>j} \PP^{(m)}[C_2, s_1, s_2, c | \calD] \qquad \text{if\,} c\neq l,c\neq r,
\end{align*}
and $\omega^{(m)}_{j,r} = 1 - \omega^{(m)}_{j,l} - \sum_c \omega^{(m)}_{j,c}$. 
%
We calculate these weights during the E-step and simply apply the negative cross-entropy loss during the M-steps to perform our EM-Gen. The initialization and training are the same as  algorithm~\ref{algo:em}.

\subsection{Prior Distribution}
\label{subsec:prior}
\noindent We apply a prior on the position of the action boundaries $s_k$. Like previous works~\cite{weakly-souri2019fast,prevsegmentrnn-kuehne2018hybrid,prevsegment-statis-richard2016temporal}, we assume that the length of an action segment for action $c$ is a random variable following a \textit{Poisson} distribution $\text{Pois}(\mu_c)$ with a mean length $\mu_c > 0$. Using this assumption, one can arrive that $\pi(s_{k}=j)$ of Eq.~\ref{eq:posteriorBoundarySF} (probability of the $k^{th}$ action boundary $s_{k}$ is located at frame $j$), follows a binomial distribution,~i.e.,~$s_k \sim \text{Bin}(T, p_{k})$, with parameters $T$, the video length, and $p_{k} = \frac{\sum_{n\le k-1} \mu_{c_n} }{ \sum_{n\le K} \mu_{c_n}}$, where $K$ is the number of action segments in the video. For EM-Gen, the prior is less straightforward and we defer it to the \textit{Supplementary}.

\subsection{Loss Function}

\noindent We apply $\calL_{\text{EM}} := -\frac{1}{T} \calQ(\param{}, \param^{(m)})$ during the M-step,~i.e.,~we minimize the negative $\calQ$-function normalized by the number of frames $T$ as a loss for learning the TCN. 
We also add two auxiliary losses, similar to~\cite{SingleFrameJGall}: a \textit{transition loss} $\calL_{\text{TR}} = \frac{1}{TC} \sum_{t,c} \min\BK{|\delta_{t,c}|,\, \epsilon}$ to penalize small inter-frame differences (up to a threshold $\epsilon$) to limit over-segmentation; 
and a \textit{confidence loss} $\calL_{\text{Conf}} = \frac{1}{T} \sum_{k=1}^K \sum_{t=t_{k-1}}^{t_{k}-1} \BK{ \delta^+_{t,y[t_{k-1}]} + \delta^-_{t,y[t_{k}]} }$ to ensure peak confidence at the annotated timestamps. Here, $\delta_{t,c} := \log p_{t,c}^{\param} - \log p_{t-1,c}^{\param}, \delta^+ := \max(\delta, 0), \delta^-:= \max(-\delta, 0)$, $C$ is the number of action classes, and $K$ is the total number of annotated timestamps in the video. It is worth noting that our posterior weights from Eq.~\ref{eq:Qfunc-as-CE} already induce this property. The final loss function is 
\begin{align*}
    \calL &= \calL_{\text{EM}} + \lambda_{\text{TR}} \cdot \calL_{\text{TR}} + \lambda_{\text{Conf}} \cdot \calL_{\text{Conf}}
\end{align*}
We use the values of $\lambda_{\text{TR}}, \lambda_{\text{Conf}}$ as suggested by \cite{SingleFrameJGall}.

\subsection{SkipTag Supervision}\label{subsec:skiptag}
Even though TSS requires annotation of very few frames, the constraint of exactly one frame for every action segment incurs significant amounts of annotation time (see Sec.~\ref{subsec:analysis-annotation}).  As the EM-Gen formulation tolerates missing frames between segments, we are motivated to introduce a more flexible form of frame-wise annotation. To this end, we propose the new \randomframe{}.
\randomframe{} is less restrictive and allows annotators to simply label the actions of a random set of timestamps placed somewhat evenly. However, two consecutively annotated timestamps $t_{k-1}$ and $t_{k}$ now have an unknown number of action boundaries in between, including no boundaries. As an experimental setting, SkipTag annotations are more challenging than TSS not only because of removing the TSS annotation constraint but also because the number of annotated timestamps do not correspond to the (unknown) number of action segments. TSS implicitly gives the number of segments per video since the annotations match the exact number of action segments. Thus, Skiptag annotation is a weaker form of supervision than TSS annotation.

To accommodate SkipTag annotations, EM-Gen must also handle the case of no action boundaries between timestamps $t_{k-1}, t_{k}$. We defer the formulation related to this new case to the \textit{Supplementary}. To limit complexity, we also do not consider beyond two action boundaries between consecutive timestamps $t_{k-1}, t_{k}$. Experimentally, this occurs very infrequently (see \textit{Supplementary}). However, this simplification is our design choice and is not a limitation of the EM-Gen framework, i.e., one can enumerate more cases in line with the same foundation. 

\section{Experiments}
\label{sec:experiments}

\textbf{Datasets:} We show our results on standard cross-validation splits of Breakfast~\cite{kuehne2014language} (1.7k, 3.6 mins/video, 48 actions), 50Salads~\cite{stein2013combining} (50, 6.4 mins/video, 19 actions) and GTEA~\cite{GTEA-fathi2011learning} (28, 1.5 mins/video, 11 actions). The standard evaluation criteria are the Mean-over-Frames (MoF), segment-wise edit distance (Edit), and $F1$-scores with IoU thresholds of $0.25$ and $0.50$ ($F1@\{25, 50\}$) (see \cite{li2020ms}).

\textbf{Implementation Details:}\label{subsec:implementation} We compare using the same modified MS-TCN as~\cite{SingleFrameJGall} as our base network. We report results on additional base networks in the \textit{Supplementary}. To initialize $\param^{(0)}$, we train the base TCN for \{30,50,50\} epochs ($N^{init}$) for Breakfast, 50Salads, and GTEA, respectively.  
Subsequently, we apply \{10,20,20\} E-M iterations ($M$), fixing each M-step to 5 epochs ($N^{max}$) for all datasets to update $\param^{(m)}$. We use the Adam optimizer~\cite{kingma2014adam} during initialization; for each M-step we use a learning rate 5e-4 and a batch-size of 8. 

\textbf{Random Seeds \& Variation in Annotations:}
All our results, including baselines, are reported for three seeds used for random selection of the annotated frames (timestamps) for both modes of supervision. Since~\cite{SingleFrameJGall} reports a single run, we report the results from that same set of timestamps in Table~\ref{tab:SOTA-comparison}.

\textbf{Baselines:}
We compare with two baselines: a `\textit{Naive}' baseline trained on \textit{only} the annotated frames (see initialization in Sec.~\ref{subsec:singleframe}) and a `\textit{Uniform}' baseline (named according to~\cite{SingleFrameJGall}, applicable only to TSS) that fixes the action boundary $s_k$ to the mid-point between consecutive timestamps $t_{k-1}, t_k$.
The action classes of unlabelled frames are then set accordingly and used during training.  

\begin{table}[t]
\caption{\textbf{EM-TSS Posterior performance.} The obtained boundaries are highly accurate with high MoF and low boundary error (\textit{`Error \%'}).}\label{tab:boundary-error-TSS}
\centering
\small
\begin{tabular}{c|cc|cc}
\hline
\multirow{2}{*}{\textbf{Method}} & \multicolumn{2}{c|}{50Salads} & \multicolumn{2}{c}{Breakfast}\\
& Error(\%) & MoF & Error(\%) & MoF \\
\hline
\hline
Uniform & 1.23 & 76.3 & 4.69 & 69.2 \\
EM-TSS & 0.55 & 89.6 & 2.19 & 87.8 \\
\hline
\end{tabular}
\end{table}

\subsection{Timestamp Supervision Results}\label{subsec:results-TSS}

\noindent \textbf{Evaluation of Formulated Posterior.} Table~\ref{tab:boundary-error-TSS} compares the performance of our boundaries estimated from EM-TSS posterior probabilities against `\textit{Uniform}' (boundary at mid-point) for all training data. The action boundary locations $s_k$ (unknown under~\singleframe{}) are obtained by taking the expectation with respect to the posterior probabilities in Eq. \ref{eq:posteriorBoundarySF}. We evaluate \textit{duration normalized} boundary error (Error (\%)) as $|\widehat{s}_k - s_k^{gt}| / T$, where $\widehat{s}_k$ and $s_k^{gt}$ are the expected and ground-truth boundary locations, respectively, and $T$ is the video duration.  
We also evaluate the frame-wise posterior weights $\omega^{(M)}_{j,c}$ (obtained after the final E-step) from Eq.~\ref{eq:Qfunc-as-CE} by regarding them as probability predictions and computing the frame-wise accuracy (MoF). Our boundary errors are less than half that of the uniform baseline; for 50Salads, our EM-TSS's boundary error is less than 1\%, while the MoF reaches 87\% or higher for Breakfast and 50Salads.

\begin{table}[h]
\caption{\textbf{Timestamp Supervision.} Our EM-TSS exceeds Fully-supervised (100\% labels) on several metrics on 50Salads and GTEA. On Breakfast and 50Salads, EM-TSS has only 0.9\% and 0.5\% less MoF than Fully-supervised. We report the average results over three randomly sampled sets of annotations.}\label{tab:single_frame_supervision}
\centering
\small
\begin{tabular}{l | cccc | cccc | cccc}
\hline
& \multicolumn{4}{c|}{50Salads} & \multicolumn{4}{c|}{Breakfast} & \multicolumn{4}{c}{GTEA} \\
\hline\hline
\textbf{Method} & \multicolumn{2}{c}{$F1\{25,50\}$} & Edit & MoF & 
                \multicolumn{2}{c}{$F1\{25,50\}$} & Edit & MoF & 
                \multicolumn{2}{c}{$F1\{25,50\}$} & Edit & MoF\\
\hline
Naive & 39.4 & 29.3 & 34.2 & 69.9 & 
        27.6 & 19.5 & 35.4 & 58.0 & 
        52.4 & 37.5 & 50.0 & 55.3\\
Uniform & 58.2 & 42.3 & 60.4 & 63.4
        & 56.3 & 36.4 & 68.1 & 51.0
        & 72.5 & 50.9 & 73.1 & 56.5 \\
EM-TSS & \textbf{75.4} & \textbf{63.7} & \textbf{70.9} & 77.3 & 
        63.0 & 49.9 & 67.0 & 67.1 & 
        81.3 & 66.4 & \textbf{81.2} & 68.6\\
\hline
Fully-supervised & 
        67.7 & 58.6 & 63.8 & \textbf{77.8} & 
        \textbf{64.2} & \textbf{51.5} & \textbf{69.4} & \textbf{68.0} & 
        \textbf{82.7} & \textbf{69.6} & 79.6 & \textbf{76.1}\\
\hline
\end{tabular}
\vspace{-1.0mm}
\end{table}

\begin{table}[h]
\caption{\textbf{Comparison with SOTA:} In Timestamp Supervision (TSS), we exceed the previous SOTA by a considerable margin for almost all metrics. Our TSS results are reported for the same timestamp set as \cite{SingleFrameJGall,SingleFrameDimaDamen}.}
\label{tab:SOTA-comparison}
\centering
\small
\begin{tabular}{l|cccc|cccc|cccc}
\hline
& \multicolumn{4}{c|}{50Salads} & \multicolumn{4}{c|}{Breakfast} & \multicolumn{4}{c}{GTEA} \\
\hline\hline
\textbf{Method} 
                & \multicolumn{2}{c}{$F1\{25,50\}$} & Edit & MoF & 
                \multicolumn{2}{c}{$F1\{25,50\}$} & Edit & MoF & 
                \multicolumn{2}{c}{$F1\{25,50\}$} & Edit & MoF\\
\hline
Plateau \cite{SingleFrameDimaDamen} & 
        68.2 & 56.1 & 62.6 & 73.9 & 
        59.1 & 43.2 & 65.9 & 63.5 & 
        68.0 & 43.6 & 72.3 & 52.9\\
Li et al. \cite{SingleFrameJGall} &
        70.9 & 60.1 & 66.8 & 75.6 &
        63.6 & 47.4 & \textbf{69.9} & 64.1 &
        73.0 & 55.4 & 72.3 & 66.4\\
\hline
\textbf{Our EM} & \textbf{75.9} & \textbf{64.7} & \textbf{71.6} & \textbf{77.9} &
        \textbf{63.7} & \textbf{49.8} & 67.2 & \textbf{67.0} &
        \textbf{82.7} & \textbf{66.5} & \textbf{82.3} & \textbf{70.5}\\
\textit{(Inc. SOTA)} & +5.0 & +4.6 & +4.8 & +2.3 
         & +0.1 & +2.4 & -2.7 & +2.9
         & +9.7 & +11.1 & +10 & +4.1 \\

\hline
Fully-supervised & 
        67.7 & 58.6 & 63.8 & 77.8 & 
        64.2 & 51.5 & 69.4 & 68.0 & 
        82.7 & 69.6 & 79.6 & 76.1\\
\hline
\end{tabular}
\vspace{-2.0mm}
\end{table}

\textbf{Comparison to Baseline and Full Supervision.} Table \ref{tab:single_frame_supervision} compares our EM-TSS with the `\textit{Naive}' and `\textit{Uniform}' baselines (see section \ref{sec:experiments}) and a fully-supervised setup. Our EM-TSS outperforms all baselines and even the fully-supervised method in several metrics for the 50Salads and GTEA datasets. Most notable is the +7\% difference in the F1\{50,25\} and Edit scores in 50Salads.

We discuss possible reasons behind EM-TSS exceeding full-supervision in the \textit{Supplementary}, and in subsection \ref{subsec:analysis-annotation}, where the (negative) effect of boundary ambiguity is shown to be more prominent in 50Salads and GTEA than Breakfast. 

\textbf{\textit{SOTA} comparison.} In table \ref{tab:SOTA-comparison}, we compare our EM-TSS with~\cite{SingleFrameDimaDamen,SingleFrameJGall}; we use the \textit{same} set of annotated timestamps as \cite{SingleFrameDimaDamen,SingleFrameJGall}.
Our method outperforms the previous SOTA for most metrics on all datasets, with margins of +11\%, +10\%, +5\% in GTEA F1@50, GTEA Edit, and 50Salads F1@25, respectively.

\subsection{Performance With Missed Segments}
\begin{figure}[h]
\begin{center}
    \includegraphics[width=1.0\linewidth]{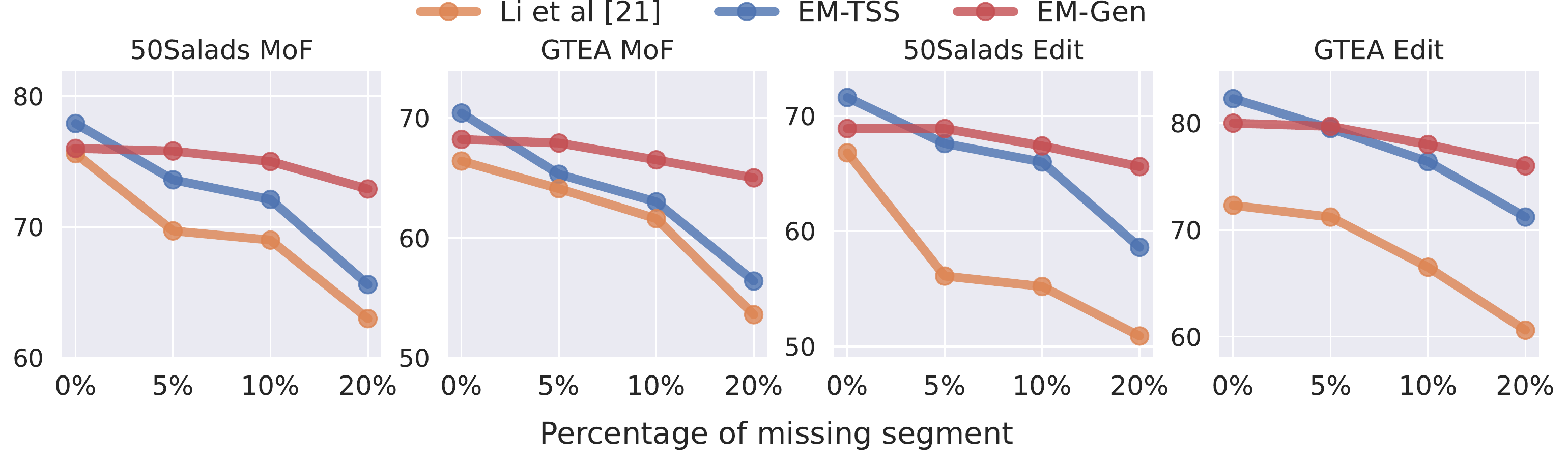}
\end{center}
\vspace{-4mm}
\caption{\textbf{Performance under (\% of) missing action segments.} Our EM-TSS performs significantly better than \cite{SingleFrameJGall} under missing action segments (0\% denotes no missed segment). Our EM-Gen is robust to missing segments, and with 20\% missing segments, EM-Gen significantly outperforms TSS methods.}
\label{fig:perf-vs-error}
\vspace{-2mm}
\end{figure}

%
In figure \ref{fig:perf-vs-error}, we show that existing TSS methods suffer from poor performance with missing segments, with a sharp drop in performance at 5\% and at 20\% missing segments.  While EM-TSS fares better, it does not exhibit the robustness of EM-Gen, as EM-Gen explicitly models missing segments between two consecutive timestamps. We detail the numerical performance in the \textit{Supplementary}.

\begin{table}[h]
\caption{\textbf{\randomframe{}.} Our EM-Gen provides a significant boost over `\textit{Naive}' and ~\cite{SingleFrameJGall}'s baselines. It performs close to full supervision in 50Salads, and outperforms it in few metrics. Results are averaged over three sets of annotations. (*) indicates the results reported in ~\cite{SingleFrameJGall}.}\label{tab:rand-frame-baselines}
\centering
\small
\begin{tabular}{l | cccc | cccc | cccc}
\hline
 & \multicolumn{4}{c|}{50Salads ($\bar{K}=19$)} & \multicolumn{4}{c|}{Breakfast ($\bar{K}=6$)} & \multicolumn{4}{c}{GTEA ($\bar{K}=32$)} \\
\hline\hline
\textbf{Method} & \multicolumn{2}{c}{$F1\{25,50\}$} & Edit & MoF & 
                            \multicolumn{2}{c}{$F1\{25,50\}$} & Edit & MoF & 
                            \multicolumn{2}{c}{$F1\{25,50\}$} & Edit & MoF\\
\hline
Naive & 49.2 & 38.1 & 46.3 & 70.4 
      & 27.0 & 18.8 & 36.4 & 59.8 
      & 64.2 & 43.4 & 65.8 & 63.1 \\  
Li et al. ~\cite{SingleFrameJGall} & 54.3 & 40.1 & 54.6 & 70.7
                        & - & - & - & *61.7
                        & 69.4 & 50.0 & 68.9 & 66.0 \\ 
\textit{EM-Gen} & \textbf{68.1} & 54.9 & \textbf{64.3} & 74.4 
                & 57.3 & 45.2 & 59.9 & 64.1
                & 76.7 & 57.9 & 73.5 & 69.8 \\
\hline
Fully-supervised & 67.7 & \textbf{58.6} & 63.8 & \textbf{77.8}
                                      & \textbf{64.2} & \textbf{51.5} & \textbf{69.4} & \textbf{68.0}
                                      & \textbf{82.7} & \textbf{69.6} & \textbf{79.6} & \textbf{76.1}\\
\hline
\end{tabular}
\vspace{-3.0mm}
\end{table}
\subsection{SkipTag Supervision Results}\label{subsec:skiptag_results} 

For SkipTag annotation, we annotate ${K}$ number of timestamps per video.  We set ${K}$ to be the average number of action segments, i.e., ${K}\!=\!\{19, 6, 32\}$ for 50Salads, Breakfast and GTEA, respectively. Having ${K}$ random frames per video results in approximately the same number of annotated frames as \singleframe{}, but is much easier for the annotator.  Table~\ref{tab:rand-frame-baselines} compares the performance of previous methods to EM-Gen under \randomframe{} and shows that EM-Gen surpasses~\cite{SingleFrameJGall} across all metrics and datasets. Moreover, in 50Salads, EM-Gen is comparable to and sometimes surpasses \textit{Fully-supervised} (i.e., 100\% labels). 



%

\subsection{Additional Results} \label{subsec:analysis-annotation}

\noindent In \textit{Supplementary}, we provide ablation related to \textbf{(1)} \textbf{different priors} (see subsection~\ref{subsec:prior}) (section S1.1), \textbf{(2)} \textbf{loss-functions} (section S1.2), \textbf{(3) TCN architectures} (section S1.4), \textbf{(4)} \textbf{effect of timestamp position} (section S1.3), other hyper-parameters, and \textbf{(5)} \textbf{qualitative analysis} (section S2). 


\textbf{Annotation Efficiency User Study.} 
We perform a user study with 5 annotators labelling 50 videos of the Breakfast dataset with different types of annotations (Full, Timestamp, and SkipTag). 
The left sub-table of table \ref{tab:annotation-analysis} reports the duration-normalized annotation time,~i.e.,~the annotation time required as a fraction of the video duration. \singleframe{} requires about two-thirds of the annotation time as full supervision, while \randomframe{} requires only one-third. Other details of the user study are given in the \textit{Supplementary}.

\begin{table}[t!]
\caption{\textit{\textbf{(left)}} Labelling cost on Breakfast. Time(\%) denotes annotation cost as a fraction of video duration. \textit{\textbf{(right)}} Choosing TSS timestamps near boundaries (start or end) lowers \textit{`Naive'} initialization performance significantly (sometimes $\ge -30\%$), indicating ambiguity in labelling the boundary frames.
}\label{tab:annotation-analysis}
\parbox{.45\linewidth}{
\centering
    \begin{tabular}{l|c|c|c}
        \hline
        \textbf{Cost} & Full-sup & Timestamp & SkipTag\\
        \hline\hline
        Frames (\%) & 100 & 0.03 & 0.03 \\
        Time (\%) & 93.3 & 65.1 & 36.4 \\
        \hline
    \end{tabular}
}
\hfill
\parbox{.45\linewidth}{
\centering
    \begin{tabular}{l|c c c c}
    \hline
         Frame Type & \multicolumn{2}{c}{F1@\{25,50\}} & Edit & MoF \\
    \hline\hline
         Random & 39.4 & 29.3 & 34.2 & 69.9 \\
         Mid & 47.7 & 34.9 & 40.9 & 69.7 \\
         Start & 12.5 & 3.2 & 17.5 & 31.9 \\
         End & 12.2 & 2.7 & 21.4 & 29.1 \\
    \hline
    \end{tabular}
}
\end{table}


\textbf{Ambiguity in Boundary Frames.} The right sub-table of table \ref{tab:annotation-analysis} compares the `\textit{Naive}' baseline on the 50Salads dataset when TSS timestamps are selected from different locations of the action segment. There is a $\ge 20\%$ difference in scores when trained with boundary frames (start/end frame) versus frames selected at \textit{(Random)} or from the middle \textit{(Mid)} of the action segment. This indicates ambiguity in boundary frame annotation (also see \cite{tresspassDimaDamen,alwassel2018diagnosing}). We further explain in the \textit{Supplementary}, how EM-TSS exceeds fully-supervised results.


\subsection{Training Complexity}
\noindent The M-step of our E-M, transformed into the traditional cross-entropy loss, is extremely efficient and takes the same per-epoch time as fully-supervised. We only update our weights at the E-step once every ($N^{max}$) 5 epochs (unlike the updates at every iteration required in\cite{SingleFrameJGall,SingleFrameDimaDamen}). 
%
Our EM-TSS training time is one magnitude less than \cite{SingleFrameJGall}. Our initialization time is exactly same as \cite{SingleFrameJGall}, whereas post-initialization, on one RTX 2080 GPU, 10 epochs (50Salads) of our EM-TSS require only 5 mins vs 50 mins for \cite{SingleFrameJGall} (official code) under the same resources.

\section{Conclusion}
\label{sec:conclusion}
\noindent We first tackle the problem of timestamp supervision by introducing a novel E-M algorithm based method. Our proposed framework is general and robust enough to work with timestamp supervision, even with mistakenly missed action segments. We also propose SkipTag supervision, a novel and much weaker form of annotation than TSS. Our method seamlessly applies to all the scenarios, producing \textit{SOTA} results, and in some cases surpassing fully-supervised results. 

\section*{Acknowledgment} 
This research is supported by the National Research Foundation, Singapore under its NRF Fellowship for AI (NRF-NRFFAI1-2019-0001). Any opinions, findings and conclusions or recommendations expressed in this material are those of the author(s) and do not reflect the views of National Research Foundation, Singapore.

\appendix
\section*{Appendix}

In Sec.~\ref{sec:ablation_studies}, we show the ablation results including the impact of prior, impact of auxiliary losses, impact of timestamp annotation position, impact of segmentation architecture, and number of epochs required for each maximization step. In Sec.~\ref{sec:qualitative_analysis}, we visualize the performance of our EM-TSS. Sec.~\ref{sec:SkipTag} details the posterior and prior for EM-Gen. In Sec.~\ref{sec:user_study}, we provide some additional details of our user study.

\section{Ablation Studies}
\label{sec:ablation_studies}

\subsection{Impact of Prior}
In Sec. 4.4 of the main text, we described the use of a length prior for Timestamp supervision. We applied a Poisson prior distribution to the action segment length with parameter $\mu_c$, which is the average action length for action $c$.
We compared \textbf{(1)} a non-informative prior, i.e., equal $\mu_c$ for all actions $c$, and with \textbf{(2)} $\mu_c$ roughly estimated from a randomly sampled video. 
As shown in Table~\ref{tab:prior-ablation}, the performance of EM-TSS is fairly robust and varies marginally for the two different priors. This is unsurprising as the likelihood contributes  significantly more to the posterior than the prior, due to the large number of frames involved in the likelihood.

\begin{table}[h]
\caption{\textbf{Effect of length prior on EM-TSS}.  Results are averaged over three sets of sampled timestamps.  The change in performance due to different priors is marginal.}\label{tab:prior-ablation}
\centering
\small
\begin{tabular}{l | ccccc | ccccc }
\hline
\multirow{2}{*}{\textbf{Prior}} & \multicolumn{5}{c|}{50Salads} & \multicolumn{5}{c}{Breakfast}  \\
\cline{2-11}\cline{2-11}
 & \multicolumn{3}{c}{$F1@\{10,25,50\}$} & Edit & MoF & \multicolumn{3}{c}{$F1@\{10,25,50\}$} & Edit & MoF\\
\hline
non-informative & 77.9 & 75.0 & 63.2 & 70.5 & 77.0
                & 66.7 & 62.9 & 50.1 & 66.9 & 67.1\\
from one video & 78.4 & 75.4 & 63.7 & 70.9 & 77.3
               & 66.8 & 63.0 & 49.9 & 67.0 & 67.1\\
\hline
\end{tabular}
\end{table}

\subsection{Impact of Auxiliary Losses} Table~\ref{tab:loss-ablation} shows the effect of the additional loss terms. In the first row, we show the performance of the base loss `$\calL_{\text{CE}}$' from the TSS work \cite{SingleFrameJGall} on both the 50Salads and Breakfast datasets.  This is a cross-entropy loss derived from hard assignments of the segment boundary to the location that minimizes their handcrafted energy function. The second row shows our base E-M loss. It is evident that our base loss yields significantly better results. Additionally, we can see that adding the auxiliary losses of \cite{SingleFrameJGall} provides some benefit for the segment metrics of Edit Distance and F1-score, but has little impact on MoF. This is unlike~\cite{SingleFrameJGall}, where the auxiliary losses also improved MoF. We believe that our frame-wise weights $\omega^{(m)}_{j,c}$ naturally embed some components of transition and confidence, so these auxiliary losses are less important for improving MoF.
\begin{table}[h]
\caption{\textbf{Effect of losses on EM-TSS (same set of timestamp as \cite{SingleFrameJGall})}. Our base E-M loss based on the negative $\calQ$-function significantly outperforms the base cross-entropy loss of TSS ~\cite{SingleFrameJGall}, which uses hard boundaries from minimizing their proposed energy function. Gradually adding the different auxiliary losses to our base loss improves Edit and F1 scores, while MoF remains largely unchanged. }\label{tab:loss-ablation}
\centering
\small
\begin{tabular}{l | ccccc | ccccc }
\hline
\multirow{2}{*}{\textbf{Losses}} & \multicolumn{5}{c|}{50Salads} & \multicolumn{5}{c}{Breakfast}  \\
\cline{2-11}\cline{2-11}
 & \multicolumn{3}{c}{$F1@\{10,25,50\}$} & Edit & MoF & \multicolumn{3}{c}{$F1@\{10,25,50\}$} & Edit & MoF\\
\hline
$\calL_{\text{CE}}$ (Base loss of \cite{SingleFrameJGall}) & 65.7 & 62.6 & 50.7 & 57.7 & 72.8 & 60.3 & 52.8 & 36.7 & 64.2 & 60.2 \\
\hline
$\calL_{\text{EM}}$ (\textit{Our} base loss) & 71.8 & 69.0 & 58.7 & 62.8 & 77.4 & 64.9 & 61.2 & 48.1 & 64.9 & 67.8\\
$\calL_{\text{EM}}\!+\!\calL_{\text{TR}}$ & 75.1 & 71.7 & 60.1 & 66.2 & 76.9 & 65.4 & 61.6 & 48.2 & 65.3 & 67.0 \\
$\calL_{\text{EM}}\!+\!\calL_{\text{TR}}\!+\!\calL_{\text{Conf}}$ & 79.9 & 75.9 & 64.7 & 71.6 & 77.7 & 67.5 & 63.7 & 49.8 & 67.2 & 67.0\\
\hline
\end{tabular}
\vspace{-2.0mm}
\end{table}
\begin{table}[h]
\caption{\textbf{Effect of timestamp locations on EM-TSS.} As expected, the start frame has the worst performance because of poor initialization. The middle and random frame have similar performance.}\label{tab:frameloc-ablation}
\centering
\small
\begin{tabular}{l|c|ccccc|ccccc}
\hline
\textbf{Location of} & \multirow{2}{*}{\textbf{Method}} & \multicolumn{5}{c}{50Salads} & \multicolumn{5}{c}{Breakfast}\\
\cline{3-12}\cline{3-12}
\textbf{timestamp} &  & \multicolumn{3}{c}{$F1\{10,25,50\}$} & Edit & MoF & 
                \multicolumn{3}{c}{$F1\{10,25,50\}$} & Edit & MoF \\
\hline
\multirow{2}{*}{Start} & Naive & 21.0 & 12.5 & 3.2 & 17.5 & 31.9
                               & 17.1 & 10.4 & 3.1 & 21.3 & 30.6\\
                       & \textit{EM-TSS} & 62.9 & 50.5 & 25.0 & 63.9 & 52.0
                                & 57.3 & 46.9 & 25.0 & 61.7 & 48.5\\
\hline
\multirow{2}{*}{Centre} & Naive & 51.0 & 47.7 & 35.0 & 40.9 & 69.7
                                & 36.8 & 32.3 & 21.9 & 37.6 & 53.4\\ 
                        & \textit{EM-TSS} & 78.4 & 76.0 & 63.5 & 71.1 & 77.1
                                 & 57.3 & 46.9 & 25.0 & 61.7 & 48.5\\
\hline
\multirow{2}{*}{Random} & Naive & 44.7 & 39.4 & 29.3 & 34.2 & 69.9
                                & 31.9 & 27.6 & 19.5 & 35.4 & 58.0\\ 
                        & \textit{EM-TSS} & 78.4 & 75.4 & 63.7 & 70.9 & 77.3
                                 & 66.8 & 63.0 & 49.9 & 67.0 & 67.1\\
\hline
\end{tabular}
\end{table}
\subsection{Impact of Timestamp Positions}

In line with the previous work \cite{SingleFrameJGall}, our EM-TSS results in the main text were obtained from random timestamp annotations within the action segments. We refer to these randomly positioned timestamp annotations as `\textit{Random}'. Alternatively, we can choose the timestamps to be some specific frame of the action segments, e.g., start frame (`\textit{Start}') and centre frame (`\textit{Centre}').

In Table~\ref{tab:frameloc-ablation}, we show the variation in the results with respect to the position of the annotated timestamps. As discussed in Section~5.4 and Table~8 of the main text, the boundary annotations are highly ambiguous. Hence, the `\textit{Naive}' performance of the start frame is extremely low. 
When we applied our EM-TSS, we observed a significant  performance increase over the baseline. However, due to the poor performance of the initialization (i.e., naive baseline), the final performance of \textit{`Start'} using EM-TSS is relatively low compared to other frame locations. Choosing the centre frame yields a similar result to the randomly chosen timestamp. This is in line with our expectation as the middle frame is well within the action segment, hence free from ambiguity.

\begin{table}[h]
\caption{\textbf{Effect of segmentation architectures on EM-TSS.} We show that our algorithm performs equally well with different kinds of architectures. `\textit{Backbone from ~\cite{SingleFrameJGall}}' refers to the modified MSTCN model used as default backbone in the TSS work \cite{SingleFrameJGall} as well as our work.}\label{tab:diff-archi}
\centering
\small
\begin{tabular}{l|c|cccc|cccc|cccc}
\hline
\multirow{2}{*}{\textbf{Backbone}} & \multirow{2}{*}{\textbf{Method}} & \multicolumn{4}{c|}{50Salads} & \multicolumn{4}{c|}{Breakfast} & \multicolumn{4}{c}{GTEA} \\
\cline{3-14}\cline{3-14}
&  & \multicolumn{2}{c}{$F1\{25,50\}$} & Edit & MoF & 
                \multicolumn{2}{c}{$F1\{25,50\}$} & Edit & MoF & 
                \multicolumn{2}{c}{$F1\{25,50\}$} & Edit & MoF\\
\hline
\hline
\multirow{2}{*}{MSTCN~\cite{farha2019ms}} & Naive
                    & 38.9 & 28.2 & 34.4	& 67.3
                    & 39.5 & 30.3 & 39.3 & 55.4
                    & 54.5 & 38.0 & 49.9 & 55.4 \\

& \textit{EM-TSS}
                    & 68.9 & 57.0 & 64.4 & 75.0
                    & 61.8 & 49.1 & 64.8 & 65.0 
                    & 80.7 & 65.0 & 80.3 & 68.3 \\

\hline
MSTCN\texttt{++} & Naive 
                    & 36.8 & 25.5 & 32.2 & 64.2 
                    & 31.0 & 21.9 & 37.4 & 58.2 
                    & 56.9 & 38.8 & 53.3 & 55.8 \\

from \cite{li2020ms} & \textit{EM-TSS} 
                    & 71.4 & 58.8 & 67.5 & 74.3 
                    & 63.4 & 49.9 & 64.1 & 66.7
                    & 82.1 & 65.7 & 79.2 & 70.3 \\

\hline
Backbone & Naive & 43.3 & 34.0 & 37.2 & 69.6 &
                   29.1 & 20.1 & 37.4 & 56.8 &
                   55.3 & 39.6 & 51.1 & 56.5\\
from ~\cite{SingleFrameJGall} & \textit{EM-TSS} & 75.9 & 64.7 & 71.6 & 77.9 &
                           63.7 & 49.8 & 67.2 & 67.0 &
                           82.7 & 66.5 & 82.3 & 70.5\\
\hline
\end{tabular}
\end{table}

\subsection{Impact of Architectures}
We adopted the modified MSTCN architecture used by \cite{SingleFrameJGall} in our work. However, results in Table~\ref{tab:diff-archi} show that our EM-TSS works equally well with other architectures such as MSTCN \cite{farha2019ms} and MSTCN\texttt{++} \cite{li2020ms}.

\subsection{Impact of $N^{max}$}
We have three hyper-parameters regarding epochs, which are $N^{init}$ (initialization epochs), $N^{max}$ (epochs per maximization step) and $M$ (number of E-M iterations). Among these, $N^{init}$ and $M$ can be set based on convergence. In Table~\ref{tab:nmax-ablation}, we show an ablation of the more non-intuitive hyper-parameter $N^{max}$ on 50Salads dataset (for one set of timestamp annotations as in \cite{SingleFrameJGall}) to study its impact on the performance of EM-TSS. The table indicates that the performance is fairly stable for different values of $N^{max}$. We chose $N^{max} = 5$ for all datasets to balance runtime efficiency and stable performance.

\begin{table}[h]
\caption{\textbf{Effect of epoch-per-maximization step $N^{max}$.} The performance of EM-TSS on 50Salads dataset is fairly stable with respect to different values of $N^{max}$, for the specific set of timestamp as chosen by \cite{SingleFrameJGall}.}\label{tab:nmax-ablation}
\centering
\small
\begin{tabular}{l | ccccc}
\hline
\multicolumn{6}{c}{50Salads} \\
\hline\hline
\textbf{$N^{max}$} & \multicolumn{3}{c}{$F1@\{10,25,50\}$} & Edit & MoF\\
\hline
3 & 77.6 & 73.5 & 60.2 & 69.8 & 74.9\\
5 & 79.9 & 75.9 & 64.7 & 71.6 & 77.9\\
7 & 77.4 & 74.8 & 64.4 & 70.0 & 77.1\\
\hline
\end{tabular}
\end{table}



\section{Qualitative Analysis}
\label{sec:qualitative_analysis}
\begin{figure}[h]
\begin{center}
\includegraphics[width=0.7\linewidth]{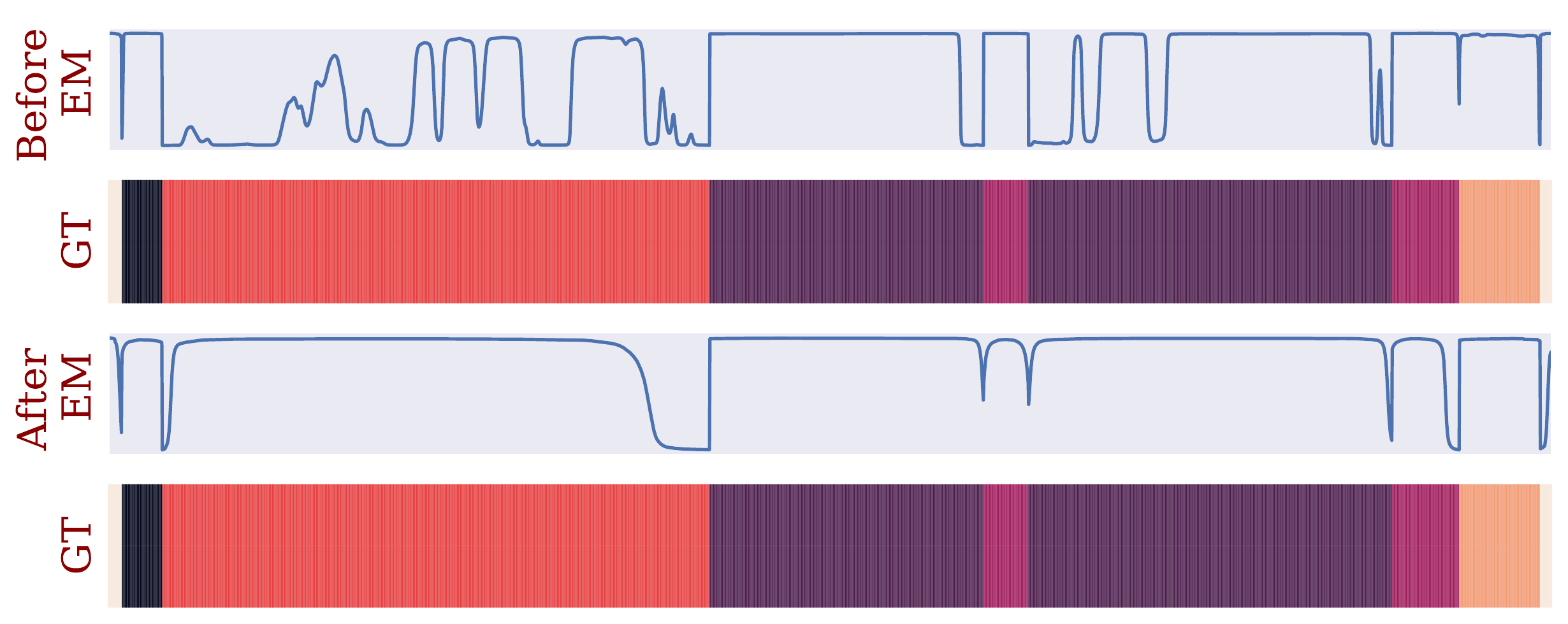}
\end{center}
\vspace{-2mm}
\caption{\textbf{Effect of EM-TSS on ground truth probability in a training video.} In the line charts, we plot the probability assigned to the correct classes. `\textit{Before EM}' denotes the probability obtained after `Naive' initialization. `\textit{After EM}' denotes the probability after convergence of EM algorithm. It is visibly clear that the probabilities are much smoother after applying E-M.}
\label{fig:before-after}
\vspace{-2mm}
\end{figure}

\begin{figure}[h]
\begin{center}
\includegraphics[width=0.7\linewidth]{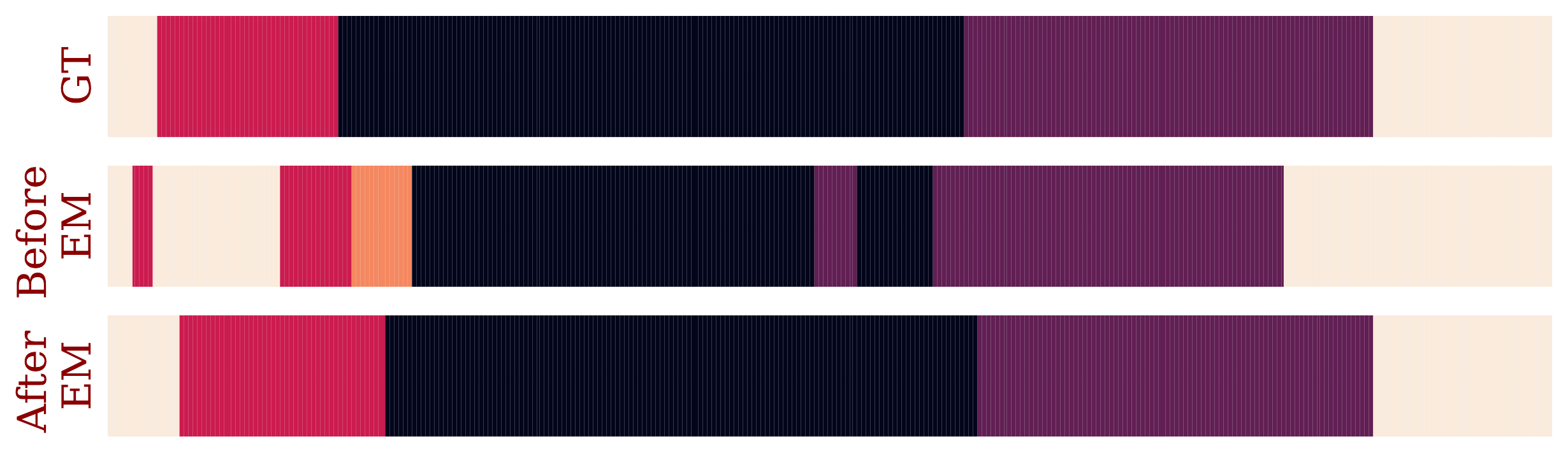}
\end{center}
\vspace{-2mm}
\caption{\textbf{Inference performance on test video.} `\textit{GT}' denotes the ground truth. `\textit{Before EM}' shows the prediction after the initialization. `\textit{After EM}' shows the prediction with the final model obtained after the convergence of the EM algorithm.}
\label{fig:before-after-test}
\vspace{-2mm}
\end{figure}

In Fig.~\ref{fig:before-after}, we focus on a specific training video and plot the class probabilities assigned to the correct classes. For example, for the entire red segment, we plot the probability assigned to that red action class in the line chart. The \textit{`GT'} plots the ground truth labels. The other two line plots show the probabilities before (obtained from `\textit{Naive}' initialization) and after the E-M algorithm (at the end of EM-TSS). It is visibly clear that the assigned probabilities improve significantly after the E-M algorithm. The naive initialization gives a highly irregular probability to the correct class. After applying E-M, the assigned probabilities are much smoother and only become uncertain near the boundaries.
Fig.~\ref{fig:before-after-test} shows the inference result on a sample test video. We show ground truth (`\textit{GT}'), predictions from the `\textit{Naive}' baseline (`\textit{Before EM}'), and predictions obtained after training with the E-M algorithm (`\textit{After EM}') in the figure. The final model prediction is considerably close to the ground truth compared to the baseline. The noticeable changes in the prediction are that there are no fragmented action segments in the prediction, and the action segments are better aligned with the ground truth segments.

\section{EM-Gen Results and Generalization to SkipTag}
\label{sec:SkipTag}

\subsection{Performance of EM-Gen Under Missing Segment} 

In Fig.~4 of the main text, we provided a comparison between EM-Gen and the TSS methods (EM-TSS and \cite{SingleFrameJGall}). Here in Table~\ref{tab:EMGEN-annotation-error}, we provide the detailed numerical results of EM-Gen under a varying degree of missing segments. Our EM-Gen outperforms the TSS work~\cite{SingleFrameJGall} when there are subtle violations in annotation constraints and remains robust even with 20\% segments missing.

\begin{table}[h]
\caption{\textbf{Performance under (\% of) missed action segments.} Our EM-TSS performs significantly better than \cite{SingleFrameJGall} under missing action segments. Our EM-Gen is robust to missing segments. With 20\% missing segments, EM-Gen significantly outperforms the TSS methods.}\label{tab:EMGEN-annotation-error}
\centering
\small
\begin{tabular}{l|c|cccc|cccc|cccc}
\hline
& & \multicolumn{4}{c|}{50Salads} & \multicolumn{4}{c|}{Breakfast} & \multicolumn{4}{c}{GTEA} \\
\hline\hline
Err\% & \textbf{Method} & \multicolumn{2}{c}{$F1\{25,50\}$} & Edit & MoF & 
                            \multicolumn{2}{c}{$F1\{25,50\}$} & Edit & MoF & 
                            \multicolumn{2}{c}{$F1\{25,50\}$} & Edit & MoF\\
\hline
\multirow{3}{*}{5\%} 
                        
 & ~\cite{SingleFrameJGall} & 
                        
                        58.5 & 44.8	& 56.1	& 69.0 &
                        57.3 & 42.0 & 60.9 & 62.4 &
                        69.7 & 53.1 & 71.2 & 64.1\\  
& \textit{EM-TSS} & 
                    
                    72.0 & 59.0 & 67.6 & 73.6 &
                    60.5 & 46.2 & 63.6 & 64.5 &
                    80.5 & 62.1 & 79.5 & 65.3\\
& \textit{EM-Gen} &
                    72.0 & 61.7 & 68.9 & 75.8 &
                    61.5 & 49.4 & 65.7 & 67.1 &
                    80.8 & 64.8 & 79.7 & 67.9\\
\hline
\multirow{3}{*}{10\%} 
& ~\cite{SingleFrameJGall} & 
                        58.1 & 43.9	& 55.2	& 69.7 &
                        53.0 & 38.2 & 58.0 & 59.5 &
                        66.3 &	51.1  & 66.5 & 61.6 \\
& \textit{EM-TSS} &
                        69.2 & 55.9 & 66.0 & 72.1 &
                        55.8 & 40.8 & 59.6 & 62.5 & 
                        77.0 & 57.3	& 76.4 & 63.0 \\
& \textit{EM-Gen} & 
                        70.9 & 58.3 & 67.4 & 75.0 & 
                        60.9 & 48.4 & 63.7 & 66.4 &
                        78.6 & 61.7 & 77.3 & 66.5\\
\hline
\multirow{3}{*}{20\%} 
& ~\cite{SingleFrameJGall} & 
                        52.2 & 36.8 & 50.9 & 63.0 & 
                        48.5 & 32.7 & 55.3 & 55.1 &
                        58.4 & 41.1 & 60.6 & 53.6\\  
& \textit{EM-TSS} & 
                    63.2 & 47.4	& 58.6 & 65.6 & 
                    50.6 & 36.6 & 57.3 & 60.5 & 
                    73.2 & 52.9 & 71.2 & 56.4 
                    \\
& \textit{EM-Gen} & 
                    68.5 & 55.7 & 66.6 & 71.9 &
                    57.9 & 45.2 & 60.1 & 65.3 &
                    75.0 & 59.7 & 74.0 & 65.0\\
\hline
\end{tabular}
\vspace{-1.0mm}
\end{table}

\subsection{Generalization of EM-Gen to SkipTag}
As briefly discussed in Sec.~4.6 of the main text, we generalized our EM-Gen method to a weaker form of annotation under SkipTag supervision. This supervision allows annotators to freely annotate a set of random timestamps spread somewhat evenly throughout the video, removing the constraint `one timestamp per action segment' of timestamp supervision. 

\begin{figure}[h]
\begin{center}
\includegraphics[width=0.4\linewidth]{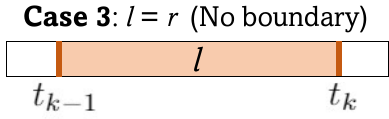}
\end{center}
\vspace{-2mm}
\caption{A typical timestamp segment with two timestamps $t_{k-1}$ and $t_{k}$ belonging to the same action segment.}
\label{fig:continuous-case}
\vspace{-2mm}
\end{figure}

We have shown that our EM-Gen is robust to one missing action segment between consecutive annotated timestamps. However, there are two other possible scenarios that EM-Gen must be generalized to. Firstly, there can be a case where two consecutive timestamps are from the same action segment, i.e., `\textit{no action boundary}' between the timestamps (depicted in Fig.~\ref{fig:continuous-case}). Secondly, there can be `\textit{more than one missing segment}' between two timestamps. Here in this section, we show how we reformulate our $\calQ$-function to take the case of `\textit{no boundary}' into account and left the case of `\textit{more than one missing segment}' out of the scope of our current work. However, the formulation can be generalized easily to handle more than two missing segments. 

\textbf{Case 3 ($C_3$): \textit{No action boundary}} between $t_{k-1}$ and $t_{k}$, as timestamp segment $\calS$ belongs to one continuous action. This case is possible only when $y[t_{k-1}] = y[t_{k}] = l$, i.e., the two action labels match, reducing the segment likelihood to
\begin{equation*}
    \calP(\calS | C_3, \param) = \prod^{t_{k}-1}_{i=t_{k-1}} p_{i,l}^{\param}.
\end{equation*}
When we take this new case into consideration along with our previously discussed cases (case 1, case 2 from Sec.4.3 of the main text), the new $\calQ$-function becomes
\begin{align*}
    \calQ(\param, \param^{(m)}) &= \sum\nolimits_s \PP^{(m)}[C_1, s | \calD] \cdot \log \calP(\calS | C_1,s, \param) \,\,\\
    &\,+\,\sum\nolimits_{\tau} \PP^{(m)}[C_2, \mathbf{\tau} | \calD] \cdot \log \calP(\calS | C_2,\tau, \param) \,\,\\
    &\,+\,\PP^{(m)}[C_3 | \calD] \cdot \log \calP(\calS | C_3, \param),  \numberthis{}\label{eq:Qfunc-randomframe2}
\end{align*}
where $\PP^{(m)}[C_3 | \calD]$ is the posterior probability that the timestamp segment $\calS$ belongs to Case 3. For SkipTag supervision, when the left and right action are not the same, i.e., $l\!\neq\!r$, the posterior weights become 
\begin{align*}
    \omega^{(m)}_{j,l} &= \sum_{s>j} \PP^{(m)}[C_1, s | \calD] + \sum_{\tau:s_1 > j}\PP^{(m)}[C_2, \tau | \calD]\\
    \omega^{(m)}_{j,c} &= \sum_{s_1\le j} \sum_{s_2>j} \PP^{(m)}[C_2, s_1, s_2, c | \calD] \qquad \text{if\,} c\neq l,c\neq r, \numberthis{\label{eq:SkipTag-qfunc}}
\end{align*}
and $\omega^{(m)}_{j,r} = 1 - \omega^{(m)}_{j,l} - \sum_c \omega^{(m)}_{j,c}$. 
When the actions are the same, $l\!=\!r$, the first term in $\omega^{(m)}_{j,l}$ is replaced by $\PP^{(m)}[C_3 | \calD]$. 

\subsection{Posterior Probabilities of EM-Gen}
In equation \ref{eq:Qfunc-randomframe}, we used the posterior probabilities to form the $\calQ$-function. In addition, we use these posterior probabilities later in equation \ref{eq:SkipTag-qfunc} to compute the per-frame posterior weights $\omega^{(m)}_{j,c}$. The form of the posterior probabilities is as follows:
\begin{align*}
    \PP^{(m)}[C_1, s \,| \calD] &= \frac{1}{Z} \cdot \calP(\calS | C_1, s, \param) \cdot \pi( C_1, s) \\
    \PP^{(m)}[C_2, \tau \,| \calD] &= \frac{1}{Z} \cdot \calP(\calS | C_2, \tau, \param) \cdot \pi( C_2, \tau) \\
    \PP^{(m)}[C_3 \,| \calD] &= \frac{1}{Z} \cdot \calP(\calS | C_3, \param) \cdot \pi( C_3)\\
    \text{where,\,\,\,\,} Z &= \sum_s  \calP(\calS | C_1, s, \param) \cdot \pi( C_1, s)  \,+ \\
    &\quad \sum_\tau  \calP(\calS | C_2, \tau, \param) \cdot \pi( C_2, \tau) \,+ \calP(\calS | C_3, \param) \cdot \pi( C_3). \numberthis{}\label{eq:posterior-randomframe}
\end{align*}

Similar to previous notations, $s$ denotes valid boundary locations of Case~1, and $\tau$ iterates over all possible triplets $(s_1, s_2, c)$ of Case~2, which are the start frame of the middle segment, start frame of the right segment and action class of the middle segment, respectively.

\subsection{Prior for EM-Gen}
Recall that for Timestamp supervision, we could directly use the Binomial prior on the location of the boundary (start) frames derived from the Poisson prior on the lengths of the action segments. This is because in TSS, we have the exact information on the (ordered) sequence of actions in the video implicitly from the timestamp annotations. For TSS with missed segments as well as SkipTag, however, we do not have this information as it is no longer guaranteed that all the action segments are annotated exactly once. 

To calculate the prior probabilities for the current timestamp segment $\calS_k$, we utilize the expected location of the \textit{last} boundary, denoted as $\beta_{k-1}$, of previous timestamp segment $\calS_{k-1}$. We naturally define $\beta_0 \equiv 0$. First, we discuss how we use $\beta_{k-1}$ to calculate priors for $\calS_k$, then we will discuss how to calculate $\beta_k$ for the segment $\calS_k$. Using $\beta_{k-1}$, we estimate the prior $\pi( C_1, s)$ for the timestamp segment $\calS_k$ as
\begin{align*}
    \pi( C_1, s ) &= \pi(s \, | C_1, \beta_{k-1}) \cdot \pi(C_1)\\
    &= \frac{e^{-\mu_l} {\mu_l}^{(s-\beta_{k-1})}}{(s-\beta_{k-1})!} \cdot \pi(C_1). \numberthis{}\label{eq:prior-case2}
\end{align*}
This is because given the last expected boundary $\beta_{k-1}$ and Case 1, the quantity $(s - \beta_{k-1})$ becomes the length of the action segment of the `\textit{left}' action class $l$ (refer to Figure 3 of main text). Hence, it follows a Poisson distribution $\text{Pois}(\mu_l)$, with $\mu_l$ being the mean length of the action class $l$. Next for Case~2, we calculate $\pi( C_2, s_1, s_2, c)$ for a triplet $(s_1, s_2, c)$, as
\begin{align*}
    \pi( C_2, s_1, s_2, c) &= \pi(s_1, s_2, c \,|  C_2, \beta_{k-1}) \cdot \pi(C_2)\\
    &= \pi(s_1 \,|  C_2, \beta_{k-1}) \cdot \pi(s_2 \,|  C_2, s_1, c) \cdot \pi(C_2,c) \\ 
    &= \frac{e^{-\mu_l} {\mu_l}^{(s_1-\beta_{k-1})}}{(s_1-\beta_{k-1})!} \cdot \frac{e^{-\mu_c} {\mu_c}^{(s_2-s_1)}}{(s_2-s_1)!} \cdot \pi(C_2,c). \numberthis{}\label{eq:prior-case3}
\end{align*}
Similar to Case 1, the quantity $(s_1 - \beta_{k-1})$ given Case 2, follows $\text{Pois}(\mu_l)$. Whereas $(s_2 - s_1) \sim \text{Pois}(\mu_c)$ when it is conditioned on Case 2 and the fact that the middle segment has action class $c$. We set the priors $\pi(C_1) = \pi(C_3) = \frac{1}{3}$. For any $c \neq l,c\neq r$, we set $\pi(C_2,c) = \frac{1}{3(C-2)}$ if $l\neq r$, and $\pi(C_2,c) = \frac{1}{3(C-1)}$ if $l=r$, where $C$ is the total number of classes. That is, we naively set all the case priors to be equally likely.

Finally, we update the last boundary as $\beta_k$ for the current timestamp segment $\calS_k$. We estimate it as
\begin{align*}
    \beta_k &= \beta_{k-1} \cdot \PP^{(m)}[C_3 \,| \calD] \,+ \sum_s  s \cdot \PP^{(m)}[C_1, s \,| \calD] \\
    &\quad + \sum_{s_2}  s_2 \cdot \sum_{s_1, c} \PP^{(m)}[C_2, s_1, s_2, c \,| \calD]. \numberthis{}\label{eq:betak}
\end{align*}
The last boundary $\beta_{k}$ for timestamp segment $\calS_k$, (1) stays the same as $\beta_{k-1}$ for Case 3 with probability $\PP^{(m)}[C_3 \,| \calD]$, (2) for Case 1 is equal to $s$ with probability $\PP^{(m)}[C_1, s \,| \calD]$, (3) and finally for Case 2 is equal to $s_2$ with probability $\sum_{s_1, c} \PP^{(m)}[C_2, s_1, s_2, c \,| \calD]$. Hence, the expected value combines all of these possible cases.

\begin{figure}[h]
    \centering
    \includegraphics[width=\linewidth]{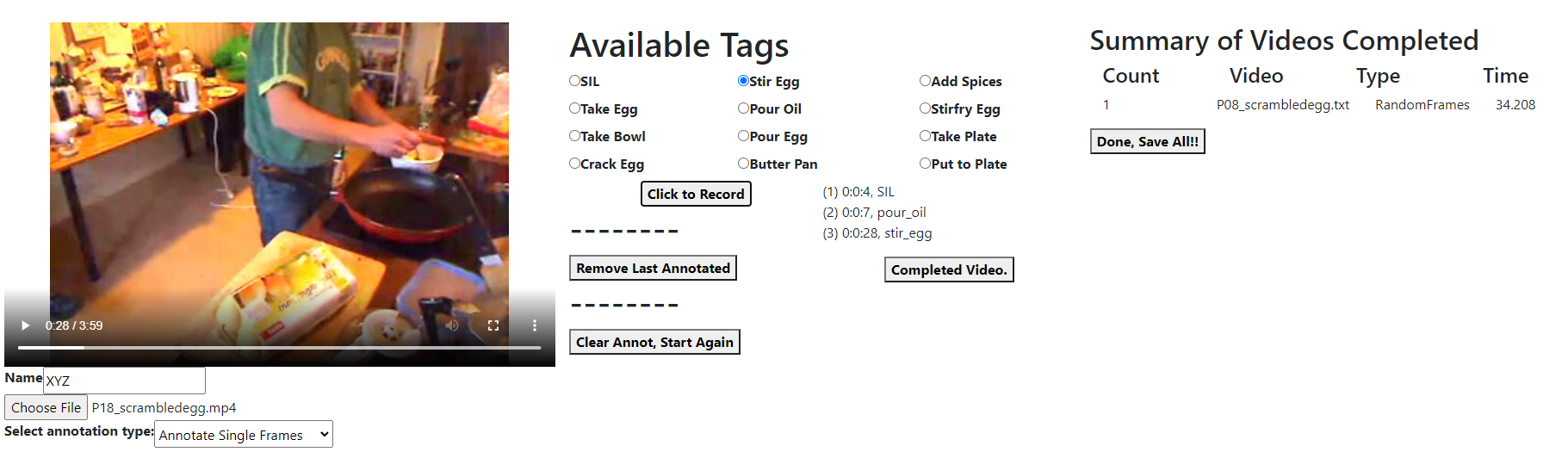}
    \caption{\textbf{User Interface for Temporal Segmentation Annotation.} Left Panel: annotators can enter id, choose video, the type of annotation and watch the video. Middle Panel: annotators can pause and select the action-tag and record it. This panel shows a list of action tags marked thus far. Right Panel: displays the tagging time summary for completed videos.}
    \label{fig:annotate_interface}
\end{figure}

\section{Details of user study} 
\label{sec:user_study}

As mentioned in Section 5.4  of the main text, our user study revealed that full-annotation (i.e., labelling the start of every action segment in the video) takes about 90\% of the video duration. Timestamp (TSS) annotation (i.e., labelling a random frame from every action segment in the video) takes about 70\% of the video duration, while SkipTag annotation (i.e., labelling 7-8 random frames in a video from Breakfast) only takes about 40\% of the video's duration. This means that for a 1-minute video, full-annotation takes about 54s, TSS annotation takes about 40s and SkipTag annotation takes about 21s on average.

A similar user study on TSS and full-annotation was conducted by ~\cite{SFNetMaEtal}. They reported that for a 1-minute video, annotation took an average of 300s for full-annotation and 50s for TSS. Though our method requires similar amount of time for TSS annotation, it reduces the full-annotation time by $5$ times compared to~\cite{SFNetMaEtal}.  We speculate that the difference is likely due to the use of different annotation interfaces. 

For full-supervision annotations, we found that the start times of the same action segments, marked by different annotators, had a standard deviation of~$\approx$~1.5 seconds or 23 frames (breakfast containing frames at 15fps). This further gives us an indication of the ambiguity in annotating the boundary frames. 

\textbf{Annotation Interface}: We developed a simple and user-friendly custom annotation interface to perform the user study for all three annotations, which is shown in ~Fig.~\ref{fig:annotate_interface}. We will make the annotation tool available along with our code repository.


%
%
\bibliographystyle{splncs04}
\bibliography{main}
\end{document}